\documentclass[10pt,twocolumn,letterpaper]{article}
\usepackage{cvpr}

\usepackage{amssymb}
\usepackage{latexsym}

\usepackage{url}
\usepackage{xcolor}
\definecolor{newcolor}{rgb}{.8,.349,.1}
\usepackage{times}
\usepackage{epsfig}
\usepackage{graphicx}
\usepackage{amsmath}
\usepackage{amssymb}
\usepackage{graphicx}
\usepackage{amsmath}
\usepackage{amssymb}
\usepackage{booktabs}
\usepackage{enumerate}
\usepackage{colortbl}
\usepackage{float}
\usepackage{algpseudocode}
\usepackage[ruled]{algorithm2e}
\usepackage{caption}
\usepackage{subcaption}
\usepackage{multirow}
\usepackage{enumitem}
\usepackage{etoolbox}
\usepackage[colorlinks]{hyperref}
\usepackage{cleveref}
\usepackage[normalem]{ulem}
\definecolor{mygray}{gray}{.9}

\DeclareMathOperator*{\argmax}{argmax}

\newcommand\dataset{Localized Audio Visual DeepFake}
\newcommand\datasetabbr{LAV-DF}
\newcommand\model{Boundary Aware Temporal Forgery Detection}
\newcommand\modelabbr{BA-TFD}
\newcommand\newmodel{Boundary Aware Temporal Forgery Detection Plus}
\newcommand\newmodelabbr{BA-TFD+}
\newcommand\task{Temporal Forgery Localization}

\newcommand\taskabbr{TFL}

\begin{document}

\setcounter{page}{1}

\title{\textit{Glitch  in the Matrix}: A Large Scale Benchmark for Content Driven Audio-Visual Forgery Detection and Localization}

\author{Zhixi Cai$^1$, Shreya Ghosh$^{2}$, Abhinav Dhall$^{3}$, Tom Gedeon$^{2}$, Kalin Stefanov$^1$, Munawar Hayat$^1$\\
$^1$Monash University, $^2$ Curtin University, $^3$ Indian Institute of Technology Ropar, \\
\small \texttt{\{zhixi.cai,kalin.stefanov,munawar.hayat\}@monash.edu,}\\ 
\small \texttt{\{shreya.ghosh,tom.gedeon\}@curtin.edu.au,abhinav@iitrpr.ac.in}}

\maketitle

\begin{abstract}
Most deepfake detection methods focus on detecting spatial and/or spatio-temporal changes in facial attributes and are centered around the binary classification task of detecting whether a video is real or fake. This is because available benchmark datasets contain mostly visual-only modifications present in the entirety of the video. However, a sophisticated deepfake may include small segments of audio or audio-visual manipulations that can completely change the meaning of the video content. To addresses this gap, we propose and benchmark a new dataset, \dataset{} (\datasetabbr{}), consisting of strategic content-driven audio, visual and audio-visual manipulations. The proposed baseline method, \model{} (\modelabbr{}), is a 3D Convolutional Neural Network-based architecture which effectively captures multimodal manipulations. We further improve (\ie \newmodelabbr{}) the baseline method by replacing the backbone with a Multiscale Vision Transformer and guide the training process with contrastive, frame classification, boundary matching and multimodal boundary matching loss functions. The quantitative analysis demonstrates the superiority of \newmodelabbr{} on temporal forgery localization and deepfake detection tasks using several benchmark datasets including our newly proposed dataset. The dataset, models and code are available at \href{https://github.com/ControlNet/LAV-DF}{https://github.com/ControlNet/LAV-DF}.

\end{abstract}


\section{Introduction}
\label{sec:introduction}

\footnote{The paper is under consideration/review at Computer Vision and Image Understanding Journal.}Increasingly powerful deep learning algorithms (\eg Autoencoders~\cite{rumelhartLearning1985} and Generative Adversarial Networks~\cite{goodfellowGenerative2020}) accompanied by the rapid advances in computing power have enabled the generation of highly realistic synthetic media commonly referred to as \textit{deepfakes}\footnote{In the text, \textit{deepfake} and \textit{forgery} are used interchangeably.}. Audio-visual deepfake content generation utilizes methods for voice cloning~\cite{wangTacotron2017, jiaTransfer2018}, face reenactment~\cite{tulyakovMoCoGAN2018, prajwalLip2020}, and face swapping~\cite{korshunovaFast2017, nirkinFsgan2019}.

Audio-visual deepfakes include videos that have been either manipulated or created from scratch to primarily mislead, deceive or influence audiences. Given that access to deepfake generation technologies has become widespread and the technologies are easy to use, some researchers argue that deepfakes are ``threat to democracy''~\cite{schwartzYou2018, brandonThere2019, sampleWhat2020, thomasDeepfakes2020}. For example,~\cite{thiesNeural2020} used a video of the former United States president Barack Obama to demonstrate a novel face reenactment method. In the resultant realistic video, the former president's lip movement is synchronized with the speech of another person. This type of manipulations has the potential to mislead people in forming wrong opinions and could have serious consequences.

Given the rapid grow of fake videos on the Internet, robust and accurate deepfake detection methods are increasingly important. This triggered the release of several benchmark datasets for deepfake detection~\cite{korshunovDeepFakes2018, rosslerFaceForensics2019, dolhanskyDeepFake2020, heForgeryNet2021} and state-of-the-art deepfake detection methods~\cite{chenDetecting2022, razaMultimodaltrace2023, ilyasAVFakeNet2023,bayarDeep2016, cozzolinoRecasting2017, yangExposing2019, liFace2020} demonstrate promising performance on those benchmark datasets, which define the problem as a binary classification task (i.e. classify the whole input video as \textit{real} or \textit{fake}).

\begin{figure}[t]
\centering
\includegraphics[width=\columnwidth]{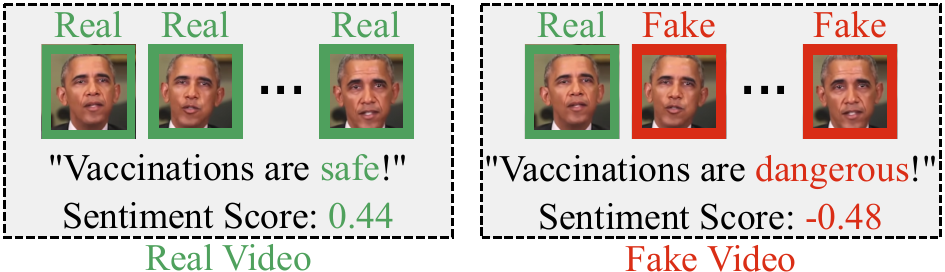}
\caption{\textbf{Content-driven audio-visual manipulation.} In the real video (left) the subject is saying ``Vaccinations are safe''. In the audio-visual deepfake (right) created from the real video, ``safe'' is changed to ``dangerous'' (resulting in a significant change in perceived sentiment). The green-edge frames are real and red-edge are fake. \textit{Note that through a subtle audio-visual manipulation, the meaning of the video content has been completely changed.}}
\label{fig:overview}
\end{figure}

Fake content however, may only constitute a small part(s) of a long real video~\cite{chughNot2020} and these modified segment(s) could completely change the meaning and sentiment of the original content. Lets consider the example illustrated in Figure~\ref{fig:overview}, where the real video on the left captures the person saying ``Vaccinations are safe''. When the word ``safe'' is replaced with its antonym ``dangerous'', the meaning and sentiment of the video is significantly changed. This type of video forgeries can effectively manipulate the public opinion, particularly when targeting media involving famous individuals, as the example with Barack Obama. Given the underlying assumption (i.e. deepfake detection is a binary classification problem) of the current deepfake detection benchmark datasets and methods, it is possible that the state-of-the-art techniques may not perform well in identifying this new type of manipulations.

This paper addresses the important task of content-driven forgery localization and detection in video. In terms of benchmark datasets, there is a significant gap in the availability of datasets for multimodal content-driven forgery localization and detection. This paper proposes a pipeline for generating such large-scale dataset that can serve as a valuable resource for future research in this area. Furthermore, this paper also introduces a novel multimodal method that utilizes audio and visual information to precisely detect the boundaries of fake segments in videos. The \textbf{main contributions} of our work are,

\begin{itemize}
\item{A large-scale public dataset, \textit{\dataset{}}, for temporal forgery localization and detection.}
\item{A multimodal method, \textit{\newmodel{}}, for fake segment localization and detection.}
\item{A thorough validation of the method's components and comprehensive comparison with the state-of-the-art.}
\end{itemize}

\begin{table*}[t]
\centering
\caption{\textbf{Details for publicly available deepfake datasets in a chronologically ascending order.} The \datasetabbr{} dataset details are reported in the last row. \textit{Cla: Classification}, \textit{SL: Spatial Localization}, \textit{\taskabbr{}: \task{}}, \textit{FS: Face Swapping}, and \textit{RE: ReEnactment}.}
\label{tab:datasets}
\scalebox{0.83}{
\begin{tabular}{l||c|c|c|c|c|c|c|c}
\toprule[0.4mm]
\rowcolor{mygray}\textbf{Dataset} & \textbf{Year} & \textbf{Tasks} & \textbf{Manipulated} & \textbf{Manipulation} & \textbf{\#Subjects} & \textbf{\#Real} & \textbf{\#Fake} & \textbf{\#Total} \\
\rowcolor{mygray}&  &  & \textbf{Modality} & \textbf{Method} &  &  &  & \\
\hline\hline
DF-TIMIT~\cite{korshunovDeepFakes2018} & 2018 & Cla & V &  FS & 43 & 320 & 640 & 960 \\
UADFV~\cite{yangExposing2019} & 2019 & Cla & V & FS & 49 & 49 & 49 & 98  \\
FaceForensics++~\cite{rosslerFaceForensics2019} & 2019 & Cla & V & FS/RE & - & 1,000 & 4,000 & 5,000 \\
Google DFD~\cite{nickContributing2019} & 2019 & Cla & V & FS & - & 363 & 3,068 & 3,431 \\
DFDC~\cite{dolhanskyDeepFake2020} & 2020 & Cla & AV & FS & 960 & 23,654 & 104,500 & 128,154 \\
DeeperForensics~\cite{jiangDeeperForensics12020} & 2020 & Cla & V & FS & 100 & 50,000 & 10,000 & 60,000 \\
Celeb-DF~\cite{liCelebDF2020} & 2020 & Cla & V & FS & 59 & 590 & 5,639 & 6,229 \\
WildDeepfake~\cite{ziWildDeepfake2020} & 2020 & Cla & - & - & - & 3,805 & 3,509 & 7,314 \\
FFIW$_{10K}$~\cite{zhouFace2021} & 2021 & Cla & V & FS & - & 10,000 & 10,000 & 20,000 \\
KoDF~\cite{kwonKoDF2021} & 2021 & Cla & V & FS/RE & 403 & 62,166 & 175,776 & 237,942 \\
FakeAVCeleb~\cite{khalidFakeAVCeleb2021} & 2021 & Cla & AV & RE & 600$+$ & 570 & 25,000$+$ & 25,500$+$ \\
ForgeryNet~\cite{heForgeryNet2021} & 2021 & SL/\taskabbr{}/Cla & V & Random FS/RE & 5,400$+$ & 99,630 & 121,617 & 221,247 \\
DF-Platter~\cite{narayanDFPlatter2023} & 2023 & Cla & V & FS & 454 & 133,260 & 132,496 & 265,756 \\
\hline
\datasetabbr~(ours) & 2022 & \taskabbr{}/Cla & AV & Content-driven RE & 153 & 36,431 & 99,873 & 136,304 \\
\bottomrule[0.4mm]
\end{tabular}}
\end{table*}

\section{Related Work}
\label{sec:related_work}
This section reviews the relevant literature on deepfake detection datasets and methods. Given the similarities between temporal forgery localization and temporal action localization, previous work in the latter area is also reviewed.

\subsection{Deepfake Detection Datasets}

Deepfake detection research is driven by datasets generated with various deepfake generation approaches. We present a summary of the deepfake detection datasets available to the research community in Table~\ref{tab:datasets}. The first deepfake dataset named DF-TIMIT was proposed by~\cite{korshunovDeepFakes2018}. DF-TIMIT curation process involved face swapping on VidTimit dataset~\cite{sandersonVidTIMIT2002}. Later, UADFV~\cite{yangExploring2018}, FaceForensics++~\cite{rosslerFaceForensics2019} and Google DFD~\cite{nickContributing2019} were introduced, and FaceForensics++ has become a popular benchmark dataset for multiple deepfake detection methods~\cite{wangM2TR2022, qianThinking2020}. The main limitation of the aforementioned datasets is their size (i.e. a maximum of thousands of video samples). Given that CNNs and Transformers (commonly used for deeepfake detection) are data-demanding techniques, these datasets have low generalization capability~\cite{liCelebDF2020}. In 2020, Facebook (\ie Meta) published the large-scale dataset DFDC~\cite{dolhanskyDeepFake2020} for deepfake detection with more than 100K samples. Until today, DFDC is the standard benchmark used for deepfake detection methods~\cite{yangAVoiDDF2023, chenDetecting2022}. After DFDC, several datasets targeting different specializations were introduced. For example, WildDeepfake~\cite{ziWildDeepfake2020} for web-crawled in-the-wild fake video detection, FFIW$_{10K}$~\cite{zhouFace2021} for detecting fake faces in videos containing multiple faces, KoDF~\cite{kwonKoDF2021} for Korean deepfake detection, and DF-Platter~\cite{narayanDFPlatter2023} for detecting multi-face heterogeneous deepfakes. DeeperForensics~\cite{jiangDeeperForensics12020} is another notable dataset that overcomes the bias of having high number of fake videos. However, all those datasets mainly consider visual-only deepfake detection. In 2021, FakeAVCeleb~\cite{khalidFakeAVCeleb2021} was introduced including both face swapping and audio-based face reenactment. This dataset includes fake audio generated from SV2TTS~\cite{jiaTransfer2018}, which makes it the first deepfake detection dataset focusing on audio-visual manipulations.

Given that all of the those datasets regard the deepfake detection as a binary classification problem, ForgeryNet~\cite{heForgeryNet2021} dataset was introduced, which includes visual-only face swapping in random frames and is suitable for both video/image classification and spatial/temporal forgery localization. However, ForgeryNet only applies random face swapping in the visual modality and does not consider audio and content-driven modifications. To bridge this gap, we propose a multimodal content-driven temporal forgery localization and detection dataset.

\subsection{Deepfake Detection Methods}

Deepfake detection methods can be categorized into two categories: traditional machine learning and deep learning approaches. The traditional machine learning methods include EM~\cite{guarneraDeepFake2020} and SVM~\cite{yangExposing2019}. On the other hand, deep learning methods include CNN~\cite{delimaDeepfake2020}, RNN~\cite{montserratDeepfakes2020, chenDetecting2022} and ViT~\cite{wodajoDeepfake2021, heoDeepFake2023, coccominiCombining2022}. Most of the prior deepfake detection methods focus on temporal inconsistencies~\cite{lewisDeepfake2020, guSpatiotemporal2021} and multimodal synchronization~\cite{chughNot2020, wangM2TR2022, mittalEmotions2020, zhuAVForensics2023} to detect fake videos.

All of the above mentioned methods employ classification centric approach. Thus, those methods do not have temporal localization capabilities. Only MDS~\cite{chughNot2020} demonstrated scenarios where only parts of the video are modified, although this approach is primarily designed for classification. Our dataset and method are designed to consider both audio-visual deepfake detection and temporal localization.

\subsection{Temporal Action Localization}

Since the temporal forgery localization task is similar to temporal action localization, we also review the literature in this domain. For temporal action localization, ActivityNet~\cite{cabaheilbronActivityNet2015}, THUMOS14~\cite{idreesTHUMOS2017}, HACS~\cite{zhaoHACS2019}, EPIC-KITCHEN~\cite{damenRescaling2022}, and FineAction~\cite{liuFineAction2022} are popular benchmark datasets. Temporal action localization methods can be classified as two types: 2-stage approaches~\cite{zengGraph2019, xuGTAD2020, liuMultiShot2021}, where the temporal bounding box proposals are generated at first and then are classified as different classes, and 1-stage approaches~\cite{linSingle2017, buchEndtoend2019, nawhalActivity2021, zhangActionFormer2022, liuEmpirical2022, liuEndtoEnd2022, shiTriDet2023, yangBasicTAD2023}, which directly predict the final temporal segments.

For temporal forgery localization, there is no requirement to classify the foreground segments, in other words, the background is always real and the foreground is always fake. Hence, 1-stage temporal action localization approaches are more relevant for the task. According to~\cite{bagchiHear2022}, these approaches can be grouped in two main categories: methods based on anchors and methods based on predicting the boundary probabilities. Anchor-based methods~\cite{shouTemporal2016, shouCDC2017, gaoTURN2017, gaoCTAP2018} utilize sliding windows in the video to detect segments. \cite{linBSN2018} proposed a new framework to generate proposals that predicts the boundary probabilities based on start and end timestamps. This approach can access the global context information to generate more precise and flexible segment proposals than anchor-based methods. Based on this method, several other approaches were proposed to enhance performance~\cite{linBMN2019, suBSN2021}.

All temporal action localization methods described above are visual-only, which is not optimal for the task of temporal forgery localization. The importance of accessing the multimodal information for temporal action localization was recently raised by~\cite{bagchiHear2022}.

\subsection{Proposed Multimodal Approach}

This paper proposes a multimodal method for precise boundary proposal estimation to detect and localize fake segments videos. We quantitatively compare the performance of the proposed method with existing state-of-the-art approaches, including BMN~\cite{linBMN2019}, AGT~\cite{nawhalActivity2021}, MDS~\cite{chughNot2020}, AVFusion~\cite{bagchiHear2022}, BSN++~\cite{suBSN2021}, TadTR~\cite{liuEndtoEnd2022}, ActionFormer~\cite{zhangActionFormer2022}, and TriDet~\cite{shiTriDet2023}.

\begin{figure*}[t]
\centering
\includegraphics[width=\textwidth]{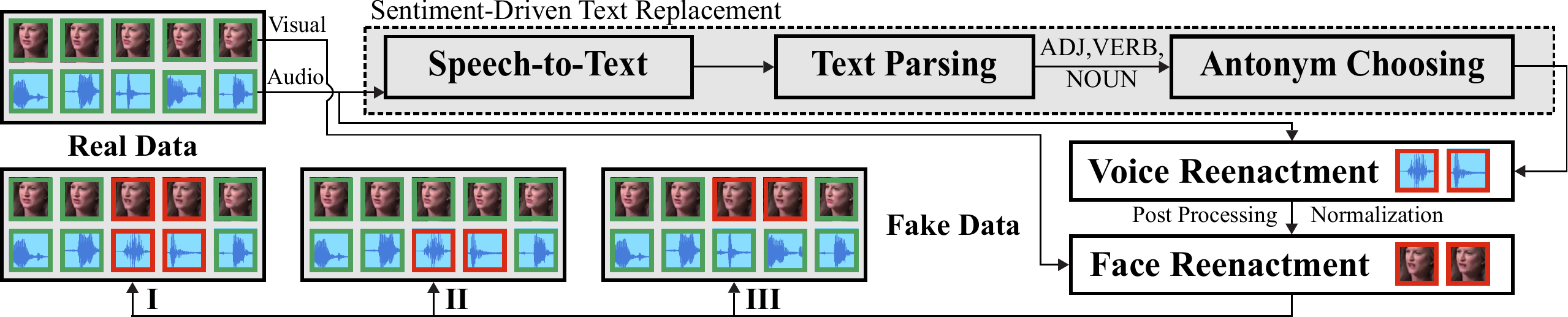}
\caption{\textbf{Content-driven audio-visual manipulation for the creation of the \datasetabbr{} dataset}. The real transcript is used to find the word tokens for replacement based on the largest change in perceived sentiment. Then the modified tokens are used as input for generating audio. Post-processing and normalization are applied to the generated audio to maintain loudness consistency in the temporal neighborhood. The generated audio is then used as input for facial reenactment. The green-edge audio and visual frames are real data, and red-edge are fake data. \textit{In total, three categories of data are generated: Fake Audio and Fake Visual, Fake Audio and Real Visual and Real Audio and Fake Visual}.}
\label{fig:data_generation}
\end{figure*}

\section{Localized Audio Visual DeepFake Dataset}
\label{sec:proposed_dataset}
We created a large-scale audio-visual deepfake dataset containing 136,304 videos (36,431 real and 99,873 fake). Our data generation pipeline is illustrated in Figure~\ref{fig:data_generation}. The generation is guided by relevant words in the video transcripts and specifically, the manipulation strategy is to replace strategic words with their antonyms, which leads to a significant change in the perceived sentiment of the statement.

\subsection{Audio-Visual Data Sourcing}

The real videos in this dataset are collected from the VoxCeleb2 dataset~\cite{chungVoxCeleb22018}, which is a large-scale facial video dataset containing more than 1 million utterance videos of 6,112 speakers. To ensure consistency, the faces within these videos are tracked and cropped using the Dlib facial detector~\cite{kingDlibml2009} at 224$\times$224 resolution. The VoxCeleb2 dataset offers a diverse range of video lengths, spoken languages, and voice qualities. Our dataset includes only English-speaking videos, where the spoken language was detected through the confidence score generated by the Google Speech-to-Text service\footnote{\url{https://cloud.google.com/speech-to-text}}. We leveraged the same service to generate the transcripts.

\subsection{Audio-Visual Data Generation}
After sourcing the real videos, the next step is to analyze each video transcript for content-driven deepfake generation. The generation process includes transcript manipulation, followed by generation of the corresponding audio and visual modalities.

\subsubsection{Transcript Manipulation}

Following the collection and wrangling of the real data, the next step is to analyze the transcript of a video denoted as $D = \{d_0, d_1, \cdots, d_m, \cdots, d_n\} $, where $d_i$ represents individual word tokens and $n$ denotes the total number of tokens in the transcript. The objective is to identify the tokens within $D$ that should be replaced in order to achieve the maximum change in perceived sentiment. This process aims to create a modified transcript $D' = \{d_0, d_1, \cdots, d_m', \cdots, d_n\}$, which consists of most of the original tokens from $D$ and the replacements for a few specific tokens. The replacement tokens, denoted as $d'$, are selected from a set $\hat{d}$ containing antonyms of $d$, sourced from WordNet~\cite{fellbaumWordNet1998}. To determine the sentiment value of the transcript, we employed the sentiment analyzer available in NLTK~\cite{birdNatural2009}.
Specifically, for each token $d$ in a transcript $D$, the replacement is found with,
$$\tau = \argmax_{d\in D, d' \in \hat{d}}|S(D)-S(D')| $$
Then all replacements in a transcript $D$ are found as follows,
$$\theta = \argmax_{\{\tau_{m}\}_{m=1}^{M}}|\sum_{i=1}^{M}\Delta S(\tau_{i})|$$
where $\Delta S(\tau_{i})$ is the difference in sentiment score of the original and modified transcripts when utilizing the replacement $\tau_{i}$ and $M$ is the maximum number of replacements.

There is up to 1 replacement for videos shorter than 10 seconds;
otherwise, there can be a maximum of 2 replacements. The shift in sentiment distribution following the manipulations is visualized in Figure~\ref{fig:data_distribution}~(a), while the histogram of $|\Delta S|$ indicating that the sentiment of most transcripts has been successfully changed, is shown in Figure~\ref{fig:data_distribution}~(b).

\subsubsection{Audio Generation}
After the transcript manipulation, the next step is to generate speaker-specific audio for the replacement tokens. Motivated by the prior work on adaptive text-to-speech methods~\cite{jiaTransfer2018, casanovaSCGlowTTS2021, neekharaExpressive2021}, we adopted SV2TTS~\cite{jiaTransfer2018} for speaker-specific audio generation.
SV2TTS consists of three modules: 1) An encoder module responsible for extracting the style embedding of the reference speaker, 2) A spectrogram generation module based on Tacotron 2~\cite{shenNatural2018} utilizing replacement tokens and the speaker style embedding, and 3) A vocoder module based on WaveNet~\cite{oordWaveNet2016}, which generates realistic audio using the spectrogram. In the audio generation, we utilized a pre-trained SV2TTS model to generate the audio segments. Then, we performed loudness normalization on the generated audio segments by considering the corresponding real audio neighbors. The rationale behind the loudness normalization is to generate a more realistic counterpart of the audio segment chosen for replacement.

\subsubsection{Video Generation}
The generated audio is used as input for generating the corresponding visual frames. Wav2Lip~\cite{prajwalLip2020} facial reenactment is used for this task, as it has been shown to achieve state-of-the-art output generation quality along with better generalization~\cite{jamaludinYou2019, krAutomatic2019}. 
We encountered several issues with using other popular visual generation methods such as AD-NeRF~\cite{guoADNeRF2021} and ATVGnet~\cite{chenHierarchical2019}. For example,
AD-NeRF does not fit in our generation context (\ie zero-shot generation of unseen speakers), and ATVGnet uses a static reference image as input for facial reenactment, resulting in pose inconsistencies on the boundaries between real and fake segments. In contrast, Wav2Lip uses a reference video and target audio as input and generates an output video in which the person in the reference video lip-syncs to the target audio content, ensuring pose consistency between real and fake segments. We employed a pre-trained Wav2Lip model and upscaled the generated visual segments to a resolution of $224\times 224$. The generated audio-visual segments are then synchronized and used to replace the original audio-visual segments.

Similar to~\cite{khalidEvaluation2021}, \datasetabbr{} includes three categories of generated data,

\begin{itemize}
\item{\textbf{Fake Audio} and \textbf{Fake Visual.} Both the real audio and visual segments corresponding to the replacement tokens are manipulated.}
\item{\textbf{Fake Audio} and \textbf{Real Visual.} Only the real audio segments corresponding to the replacement tokens are manipulated. To keep the fake audio and real visual segments synchronized, the corresponding real visual segments are length-normalized.}
\item{\textbf{Real Audio} and \textbf{Fake Visual.} Only the real visual segments corresponding to the replacement tokens are manipulated and the length of the fake visual segments is normalized to match the length of the real audio segments.}
\end{itemize}

\subsection{Dataset Statistics}

\begin{figure}[h]
\centering
\includegraphics[width=\columnwidth]{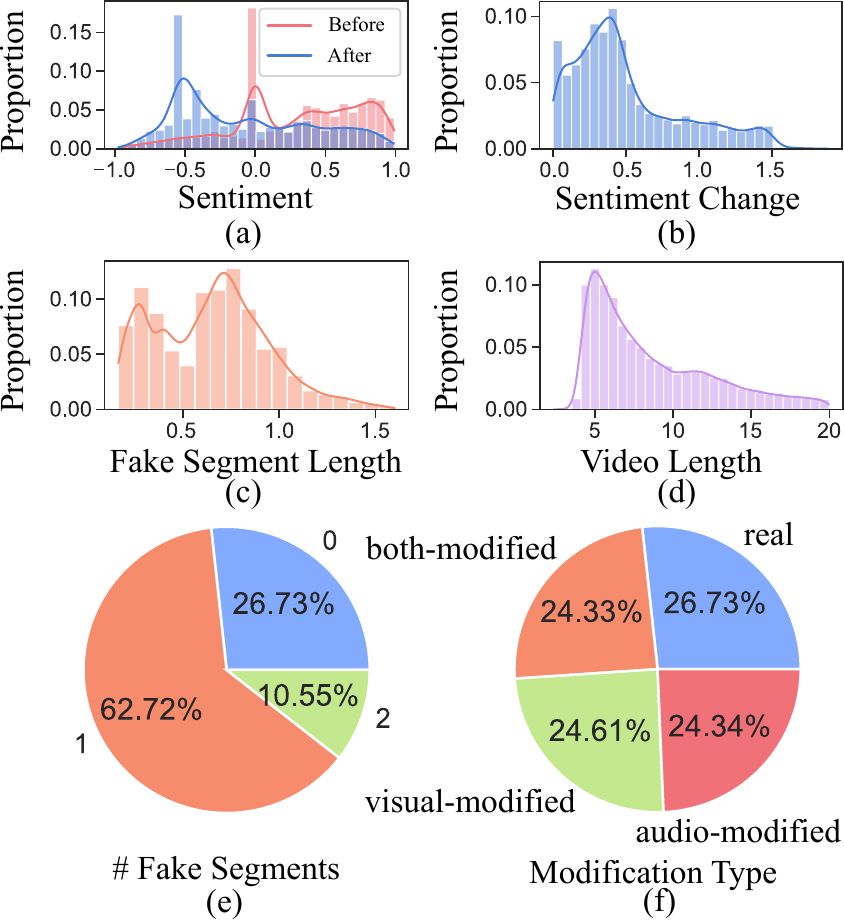}
\caption{\textbf{Statistics of the \datasetabbr{} dataset.} (a) Distribution of sentiment scores before and after content-driven deepfake generation, (b) Histogram of sentiment changes $|\Delta S|$, (c) Distribution of fake segment lengths, (d) Distribution of video lengths, (e) Proportion of number of fake segments, and (f) Proportion of modifications.}
\label{fig:data_distribution}
\end{figure}

The dataset contains 136,304 videos of 153 unique identities, with 36,431 real videos and 99,873 videos containing fake segments. For benchmarking, we splitted the dataset into 3 identity-independent subsets: train (78,703 videos of 91 identities), validation (31,501 videos of 31 identities), and test (26,100 videos of 31 identities). Summary of main statistics of the dataset is presented in Figure~\ref{fig:data_distribution}.

The dataset includes a total of 114,253 fake segments, with duration $(0,1.6]$ seconds and an average length of 0.65 seconds. Notably, 89.26\% of the fake segments are shorter than 1 second. The maximum length of the videos in the dataset is 20 seconds and 69.61\% of the videos are shorter than 10 seconds. In terms of modality modification, the distribution is balanced among the four types: visual-modified, audio-modified, both-modified and real. Additionally, the majority of the videos (62.72\%) contain only 1 fake segment, while a smaller proportion of videos (10.55\%) include 2 fake segments.

\subsection{Dataset Quality}
Table~\ref{tab:visual_quality} provides a quantitative comparison (PSNR and SSIM) with existing dataset generation pipelines in terms of visual quality, demonstrating that our pipeline achieves better visual quality on the VoxCeleb2 dataset.

\begin{table}[h]
\centering
\caption{\textbf{Visual quality of the \datasetabbr{} dataset.} We maintained the experimental protocol and adopted the scores on VoxCeleb2 for the related deepfake generation pipelines from~\cite{zhouPoseControllable2021}.}
\scalebox{1}{
\begin{tabular}{l|cc}
\toprule[0.4mm]
\rowcolor{mygray} \textbf{Method} & \textbf{PSNR} & \textbf{SSIM} \\ \hline \hline
ATVGnet~\cite{chenHierarchical2019} & 29.41 & 0.826 \\
Wav2Lip~\cite{prajwalLip2020} & 29.54 & 0.846 \\
MakeitTalk~\cite{zhouMakeltTalk2020} & 29.51 & 0.817 \\
Rhythmic Head~\cite{chenTalkingHead2020} & 29.55 & 0.779 \\
PC-AVS~\cite{zhouPoseControllable2021} & 29.68 & 0.886 \\ \hline
\datasetabbr{} (Ours) & \textbf{33.06} & \textbf{0.898} \\
\bottomrule[0.4mm]
\end{tabular}}
\label{tab:visual_quality}
\end{table}

\section{Boundary Aware Temporal Forgery Detection+ Method}
\label{sec:proposed_method}
\begin{figure*}[t]
\centering
\includegraphics[width=\textwidth]{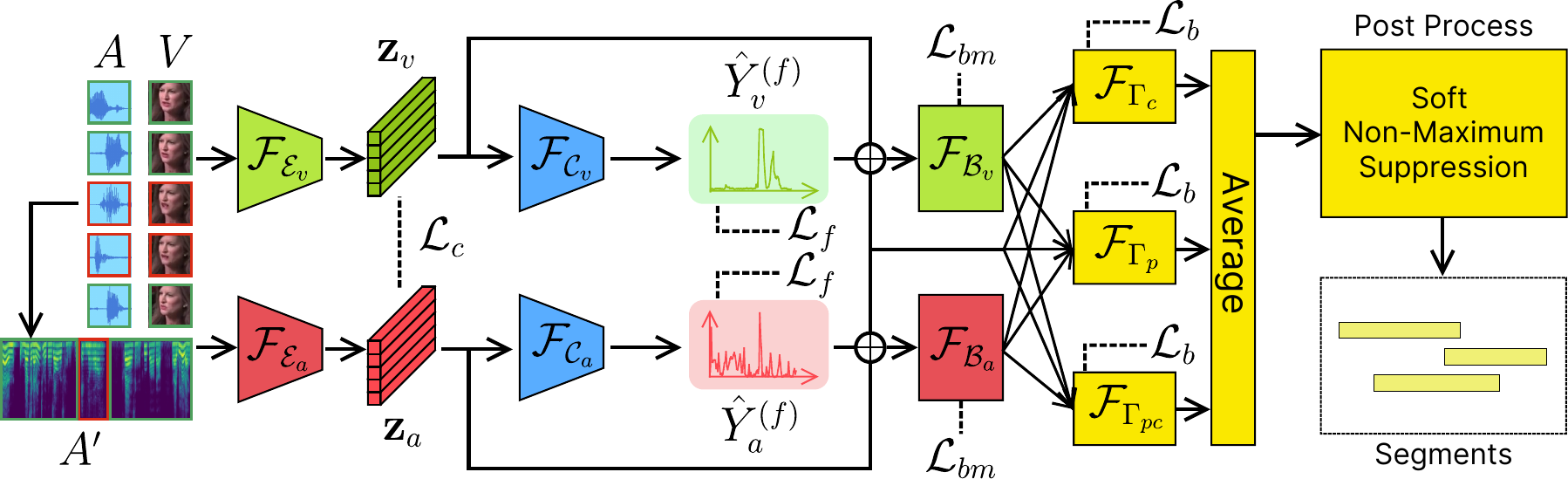}
\caption{\textbf{Overview of the \newmodelabbr{} method.} \newmodelabbr{} mainly comprises of 1) Visual encoder ($\mathcal{F}_{\mathcal{E}_v}$) that takes resized raw video frames as input, 2) Audio encoder ($\mathcal{F}_{\mathcal{E}_a}$) that takes spectrogram extracted from raw audio as input, 3) Visual and audio based frame classification module (i.e. $\mathcal{F}_{\mathcal{C}_v}$ and $\mathcal{F}_{\mathcal{C}_a}$), 4) Boundary localization module to facilitate forgery localization in both visual ($\mathcal{F}_{\mathcal{B}_v}$) and audio ($\mathcal{F}_{\mathcal{B}_a}$) modality, and finally 5) Multimodal fusion module that fuses multimodal latent features position-wise ($p$), channel-wise ($c$) and position-channel wise ($pc$). During inference, post-processing operation is applied to generate segments from the output of the fusion module. \textit{$\oplus$ denotes concatenation.}}
\label{fig:model}
\end{figure*}

The objective is to detect and localize multimodal manipulations given an input video. To this end, we designed the proposed method \newmodelabbr{} in such a way that it has the capability to capture deepfake artifacts and localize the boundary of fake segments. An overview of the proposed method is depicted in Figure~\ref{fig:model} and Algorithm~\ref{alg:train}.

    \SetKwComment{Comment}{/* }{ */}
    \begin{algorithm}[t!]
    \caption{Training procedure of \newmodelabbr}
    \label{alg:train}
    \KwData{Training data $\mathbb{D} \supset \{X_i, Y_i\}_{i=1}^{n}$, Modality modification flag $\mathbb{E} = \{\eta_i = (\eta_{vi}, \eta_{ai})\}_{i=1}^{n}$, Weights of losses $\lambda$}
    \KwResult{Parameters of the model $\theta$}
    $\theta \gets$ Initialize the parameters randomly\;
    $Y_0 \gets$ label for real data\;
    \While{$\theta$ not converged}{
        $(V, A, Y) \gets$ Next sample from $\mathbb{D}$\;
        $(\eta_v, \eta_a) \gets$ Next flag from $\mathbb{E}$\;
        $Y_v \gets$ if $\eta_v$ then $Y$ else $Y_0$\;
        $Y_a \gets$ if $\eta_a$ then $Y$ else $Y_0$\;
        $Y^{(b)} \gets$ Generate labels from $Y$\;
        $(Y^{(b)}_v, Y^{(f)}_v) \gets$ Generate labels from $Y_v$\;
        $(Y^{(b)}_a, Y^{(f)}_a) \gets$ Generate labels from $Y_a$\;
        $\mathbf{z}_v \gets \mathcal{F}_{\mathcal{E}_v}(V)$\; 
        $\mathbf{z}_a \gets \mathcal{F}_{\mathcal{E}_a}($mel-spetrogram$(A))$\;
        $Y^{(c)} \gets \eta_{vi} \land \eta_{ai}$\;
        $\mathcal{L}_c \gets~$ContrastiveLoss$(\mathbf{z}_v, \mathbf{z}_a, Y^{(c)})$\;
        $\hat{Y}^{(f)}_v \gets \mathcal{F}_{\mathcal{C}_v}(\mathbf{z}_v)$\; 
        $\hat{Y}^{(f)}_a \gets \mathcal{F}_{\mathcal{C}_a}(\mathbf{z}_a)$\;
        \Comment{FL: Frame Loss}
        $\mathcal{L}_f \gets \frac{1}{2}($FL$(\hat{Y}^{(f)}_v, Y^{(f)}_v) + $FL$(\hat{Y}^{(f)}_a, Y^{(f)}_a))$\; 
        \Comment{$\oplus$: concatenation}
        $(\hat{Y}_v^{(b)(p)}, \hat{Y}_v^{(b)(c)}, \hat{Y}_v^{(b)(pc)}) \gets \mathcal{F}_{\mathcal{B}_v}(\mathbf{z}_v \oplus \hat{Y}^{(f)}_v)$\; 
        $(\hat{Y}_a^{(b)(p)}, \hat{Y}_a^{(b)(c)}, \hat{Y}_a^{(b)(pc)}) \gets \mathcal{F}_{\mathcal{B}_a}(\mathbf{z}_a \oplus \hat{Y}^{(f)}_a)$\;
        $\hat{Y}^{(b)(p)} \gets \mathcal{F}_{\Gamma_{p}}(\hat{Y}_v^{(b)(p)}, \hat{Y}_a^{(b)(p)}, \mathbf{z}_v, \mathbf{z}_a)$\;
        $\hat{Y}^{(b)(c)} \gets \mathcal{F}_{\Gamma_{c}}(\hat{Y}_v^{(b)(c)}, \hat{Y}_a^{(b)(c)}, \mathbf{z}_v, \mathbf{z}_a)$\;
        $\hat{Y}^{(b)(pc)} \gets \mathcal{F}_{\Gamma_{pc}}(\hat{Y}_v^{(b)(pc)}, \hat{Y}_a^{(b)(pc)}, \mathbf{z}_v, \mathbf{z}_a)$\;
        $\mathcal{L}_{bm} \gets \frac{1}{2}(MSE(\hat{Y}_v^{(b)(p)}, Y^{(b)}_v) + MSE(\hat{Y}_v^{(b)(c)}, Y^{(b)}_v) + MSE(\hat{Y}_v^{(b)(pc)}, Y^{(b)}_v) + MSE(\hat{Y}_a^{(b)(p)}, Y^{(b)}_a) + MSE(\hat{Y}_a^{(b)(c)}, Y^{(b)}_a) + MSE(\hat{Y}_a^{(b)(pc)}, Y^{(b)}_a))$\;
        $\mathcal{L}_b \gets MSE(\hat{Y}^{(b)(p)}, Y^{(b)}) + MSE(\hat{Y}^{(b)(c)}, Y^{(b)}) + MSE(\hat{Y}^{(b)(pc)}, Y^{(b)})$\;
        $\theta \gets Adam(\mathcal{L}_b, \mathcal{L}_{bm}, \mathcal{L}_f, \mathcal{L}_c, \lambda, \theta)$\;
    }
    \Return $\theta$\;
    \end{algorithm}

\begin{figure}[t]
\centering
\includegraphics[width=\columnwidth]{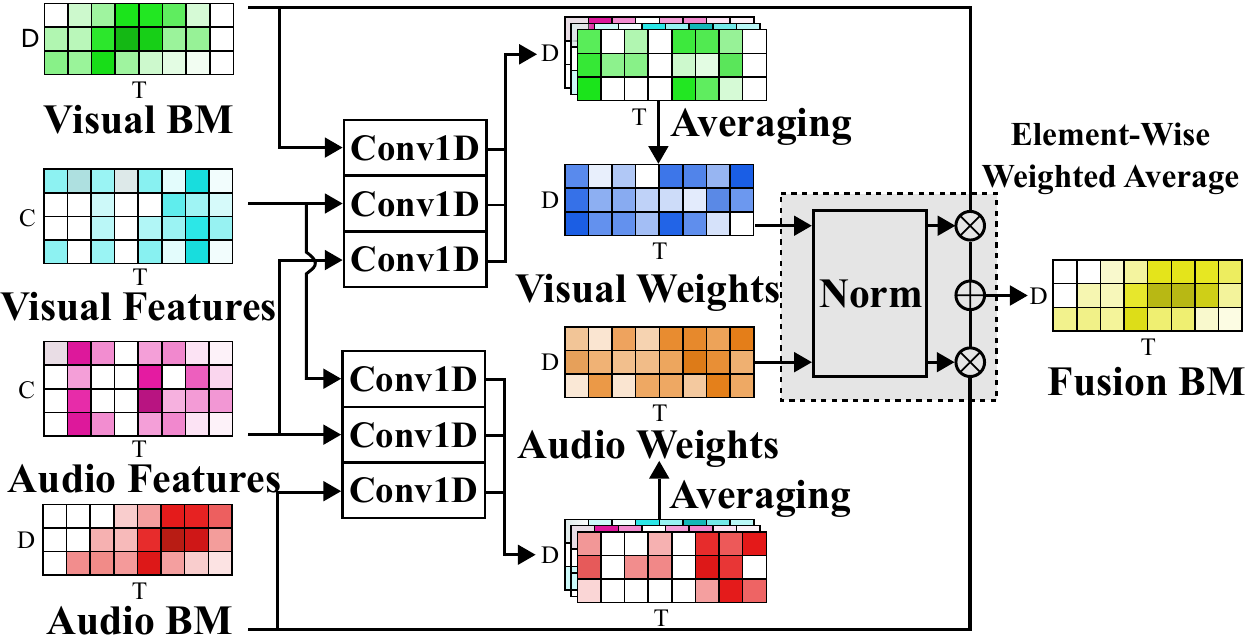}
\caption{\textbf{Overview of the \newmodelabbr{} fusion module.} The gray block is used to normalize the visual and audio weights produced by the 1D convolutional layers, followed by element-wise weighted average. \textit{$\oplus$ denotes element-wise addition, $\otimes$ denotes element-wise multiplication and BM denotes boundary map.}}
\label{fig:fusion}
\end{figure}

\subsection{Preliminaries}
The training dataset $\mathbb{D} \supset \{X_i, Y_i\}_{i=1}^{n}$ comprises of $n$ multimodal inputs $X_i$ with visual modality $V_i$ and audio modality $A_i$, and the associated output labels $Y_i$. The proposed model \newmodelabbr{} with trainable parameters $\theta$ is optimized to map the inputs $X_i$ to the outputs $Y_i$.
Each $X_i$ has a different number of frames $t_i$. In order to simplify the batch training of the model, we padded the temporal axis for all $X_i$ to $T$.

\subsection{Visual Encoder}

The goal of the visual encoder $\mathcal{F}_{\mathcal{E}_v}$ is to capture the frame-level spatio-temporal features from the input visual modality $V \supset \{V_i\}_{i=1}^{n}$ using an MViTv2~\cite{liMViTv22022}. MViTv2 achieves seminal performance gain for different video analysis tasks including video action recognition and detection. In addition, MViTv2 leverages hierarchical multi-scale features compared to the basic ViT~\cite{dosovitskiyImage2021}. Our backbone MViTv2-Base model comprises of 4 blocks and 24 multi-head self-attention layers.
As illustrated in Figure~\ref{fig:model}, the visual encoder $\mathcal{F}_{\mathcal{E}_v}$ maps the inputs $V \in \mathbb{R}^{C\times T\times H\times W}$ ($T$ is the number of frames, $C$ is the number of channels, and $H$ and $W$ are the height and width of the frames) to latent space $\mathbf{z}_v\in \mathbb{R}^{C_f \times T}$ ($C_f$ is the dimension of the features).

\subsection{Audio Encoder}

The goal of the ViT-based~\cite{dosovitskiyImage2021} audio encoder $\mathcal{F}_{\mathcal{E}_a}$ is to learn meaningful features from the raw input audio modality $A \supset \{A_i\}_{i=1}^{n}$. Following previous work~\cite{ilyasAVFakeNet2023, yangAVoiDDF2023}, we pre-process the raw audio $A$ to generate representative mel-spectrograms $A^\prime \in \mathbb{R}^{F_m\times T_a}$ ($T_a = \tau T$ is the temporal dimension and $\tau \in \mathbb{N^*}$, $\mathbb{N^*}$ denotes positive integers, and $F_m$ is the length of the mel-frequency cepstrum features). In order to keep the audio-visual synchronization, we reshape the temporal axis of the mel-spectrograms to $\tau F_m\times T$. The reshaped spectrograms $A'$ are given as input to the ViT blocks of the audio encoder $\mathcal{F}_{\mathcal{E}_a}$\footnote{We only incorporate the multi-head self-attention layers of the ViT for the audio encoder.}. The audio encoder $\mathcal{F}_{\mathcal{E}_a}$ maps the mel-spectrograms $A'$ to the latent space $\mathbf{z}_a \in \mathbb{R}^{C_f \times T}$, where $C_f$ is the features dimension.

\subsection{Frame Classification Module}
\label{sec:classification}
We further deploy frame-level classification modules on top of the visual and audio features. Let us denote the ground truth labels for visual and audio modality as $Y^{(f)}_v$ and $Y^{(f)}_a$. 
The visual classification module $\mathcal{F}_{\mathcal{C}_v}$ maps the latent visual features $\mathbf{z}_v$ to labels $\hat{Y}^{(f)}_v \in \mathbb{R}^T$.
Similarly, the audio classification module $\mathcal{F}_{\mathcal{C}_a}$ maps latent audio features $\mathbf{z}_a$ to labels $\hat{Y}^{(f)}_a \in \mathbb{R}^T$.

\subsection{Boundary Localization Module}
This module facilitates the learning of deepfake localization. Motivated by BSN++~\cite{suBSN2021}, we adopted the proposal relation block (PRB) as the framework for the boundary maps (representation of the boundary information of all densely distributed proposals). The ground truth boundary map $Y^{(b)} \in \mathbb{R}^{D\times T}$ is generated from $Y$, where $Y^{(b)}_{ij}$ is the confidence score for a segment which starts at the $j$-th frame and ends at the $(i+j)$-th frame. The PRB module contains both a position-aware attention module (captures global dependencies) and a channel-aware attention module (captures inter-dependencies between different channels). In order to achieve localization in each modality, we deploy two boundary modules, $\mathcal{F}_{\mathcal{B}_v}$ for visual and $\mathcal{F}_{\mathcal{B}_a}$ for audio modality.

The visual boundary module $\mathcal{F}_{\mathcal{B}_v}$ input consists of the concatenation of latent features $\mathbf{z}_v$ and classification outputs $\hat{Y}^{(f)}_v$, i.e $\mathbf{z}_v \oplus \hat{Y}^{(f)}_v$. $\mathcal{F}_{\mathcal{B}_v}$ predicts the position-aware boundary maps $\hat{Y}_v^{(b)(p)} \in \mathbb{R}^{D\times T}$ and the channel-aware boundary maps $\hat{Y}_v^{(b)(c)} \in \mathbb{R}^{D\times T}$ as output. These results are aggregated by a convolutional layer which outputs position-channel boundary maps denoted as $\hat{Y}_v^{(b)(pc)} \in \mathbb{R}^{D\times T}$.
Similarly, the audio boundary module $\mathcal{F}_{\mathcal{B}_a}$ input consists of the concatenation of latent features $\mathbf{z}_a$ and classification outputs $\hat{Y}^{(f)}_a$, i.e $\mathbf{z}_a \oplus \hat{Y}^{(f)}_a$. $\mathcal{F}_{\mathcal{B}_a}$ first predicts the audio position-aware boundary maps $\hat{Y}_a^{(b)(p)}$ and channel-aware boundary maps $\hat{Y}_a^{(b)(c)}$. Then $\hat{Y}_a^{(b)(p)}$ and $\hat{Y}_a^{(b)(c)}$ are aggregated to $\hat{Y}_a^{(b)(pc)}$ using a convolutional layer.

\subsection{Multimodal Fusion Module}

The fusion module illustrated in Figure~\ref{fig:fusion}, uses boundary maps $\hat{Y}_v^{(b)(p)}$, $\hat{Y}_a^{(b)(p)}$, $\hat{Y}_v^{(b)(c)}$, $\hat{Y}_a^{(b)(c)}$, $\hat{Y}_v^{(b)(pc)}$, and $\hat{Y}_a^{(b)(pc)}$ and features $\mathbf{z}_v$ and $\mathbf{z}_a$ from the visual and audio modalities as input. Since the boundary module corresponding to each modality predicts three boundary maps, there are three fusion modules for position-aware $\mathcal{F}_{\Gamma_p}$, channel-aware $\mathcal{F}_{\Gamma_c}$ and aggregated position-channel $\mathcal{F}_{\Gamma_{pc}}$ boundary maps.

For the visual modality, the visual boundary maps and features from the visual and audio modalities are used to calculate the visual weights $W_v \in \mathbb{R}^{D\times T}$. Similarly, for the audio modality, the audio boundary maps and features from both modalities are utilized to calculate the audio weights $W_a \in \mathbb{R}^{D\times T}$. The element-wise weighted average of the fusion boundary maps predictions $\hat{Y}^{(b)(p)}$, $\hat{Y}^{(b)(c)}$ and $\hat{Y}^{(b)(pc)}$ is formed in the final step. Each boundary map $\alpha \in \{p, c, pc\}$ is calculated as follows,

$$\hat{Y}^{(b)(\alpha)} = \frac{W_v \hat{Y}_v^{(b)(\alpha)} + W_a \hat{Y}_a^{(b)(\alpha)}}{W_v + W_a},$$
where all operations are element-wise.

\subsection{Loss Functions}  
\label{sec:loss}
The training process of \newmodelabbr{} is guided by contrastive ($\mathcal{L}_c$), frame classification ($\mathcal{L}_f$), boundary matching ($\mathcal{L}_b$) and multimodal boundary matching ($\mathcal{L}_{bm}$) loss functions. 

\subsubsection{Contrastive Loss}
Contrastive loss has been proven to be helpful to eliminate the misalignment between different modalities~\cite{chungOut2017, chughNot2020}. Motivated by this, \newmodelabbr{} uses the latent visual and audio features $\mathbf{z}_v$ and $\mathbf{z}_a$ of real videos as positive pairs. On the other hand, latent features $\mathbf{z}_v$ and $\mathbf{z}_a$ with at least one modified modality are considered negative pairs (\ie $Y^{(c)} = 0$). Thus, the contrastive loss minimizes the difference between the visual and audio modalities for positive pairs (\ie $Y^{(c)} = 1$) and keeps that margin larger than $\delta$ for negative pairs. The contrastive loss is defined as follows,

$$\mathcal{L}_c = \frac{1}{C_f\sum\mathbb{T}}\sum^n_{i=1} Y^{(c)}_{i}d_{i}^{2} + (1-Y^{(c)}_i)\max(\delta - d_{i}, 0)^2 $$
$$d_i = ||\mathbf{z}_{v_i} - \mathbf{z}_{a_i}||_2,$$
where, $n$ is the number of samples in the dataset, $d_i$ is the $\ell_2$ distance between visual and audio modality in the latent space, $Y^{(c)}_i$ is the label for contrastive learning and $\mathbb{T} = \{t_i\}_0^{n}$ where $\sum\mathbb{T}$ is the total number of frames in the dataset.

\subsubsection{Frame Classification Loss}
This is a standard frame level cross-entropy loss depicted as,

$$ \mathcal{L}_f = -\frac{1}{2\sum\mathbb{T}}\sum_{m\in\{a, v\}}\sum^n_{i=1}\sum^{t_i}_{j=1}H(\hat{Y}^{(f)}_{mij},Y^{(f)}_{mij}) $$
$$ H(\hat{Y}^{(f)}, Y^{(f)}) = Y^{(f)}\log{\hat{Y}^{(f)}} + (1 - Y^{(f)})\log{(1-\hat{Y}^{(f)})} $$
$$Y^{(f)}_m = \eta_m Y^{(f)} + (1 - \eta_m)Y^{(f)}_{\phi},$$
where $n$ is the number of samples in the dataset, $t_i$ is the number of frames, $m$ is the modality (\ie audio $a$ or visual $v$), $\eta_m$ specifies whether modality $m$ is manipulated or not, $Y^{(f)}_{\phi} \in 0^T$ is the label for real videos, and $\mathbb{T} = \{t_i\}_0^{n}$ where $\sum\mathbb{T}$ is the total number of frames in the dataset. This loss enforces the visual and audio encoder to learn whether a visual frame or audio sample is real or fake.

\subsubsection{Boundary Matching Loss}
Following the standard protocol~\cite{linBMN2019, suBSN2021}, we generated the ground truth boundary maps as labels for efficient training. The fusion boundary matching loss is calculated as,

$$\mathcal{L}_b = \frac{1}{3D\sum\mathbb{T}}\sum_{\alpha\in\{p,c,pc\}}\sum^n_{i=1}\sum^D_{j=1}\sum^{t_i}_{k=1}(\hat{Y}^{(b)(\alpha)}_{ijk}-Y^{(b)}_{ijk})^2,$$
where $\alpha$ is one of the boundary map types from the boundary module, $n$ is the number of samples in the dataset, $D$ is the maximum proposal duration, $t_i$ is the number of frames, and $\mathbb{T} = \{t_i\}_0^{n}$ where $\sum\mathbb{T}$ is the total number of frames in the dataset.

\begin{table*}[t]
\centering
\caption{\textbf{Temporal forgery localization results on the ``fullset'' of the \datasetabbr{} dataset.} The visual-only version of \newmodelabbr{} uses the output from the visual boundary matching layer, illustrating the performance when using only the visual modality.}
\scalebox{1}{
\begin{tabular}{l||ccc|cccc}
\toprule[0.4 mm]
\rowcolor{mygray}\textbf{Method} & \textbf{AP@0.5} & \textbf{AP@0.75} & \textbf{AP@0.95} & \textbf{AR@100} & \textbf{AR@50} & \textbf{AR@20} & \textbf{AR@10} \\ \hline \hline
BMN~\cite{linBMN2019} & 10.56 & 01.66 & 00.00 & 48.49 & 44.39 & 37.13 & 31.55 \\
BMN (E2E) & 24.01 & 07.61 & 00.07 & 53.26 & 41.24 & 31.60 & 26.93 \\
MDS~\cite{chughNot2020} & 12.78 & 01.62 & 00.00 & 37.88 & 36.71 & 34.39 & 32.15 \\
AGT~\cite{nawhalActivity2021} & 17.85 & 09.42 & 00.11 & 43.15 & 34.23 & 24.59 & 16.71 \\
BSN++~\cite{suBSN2021} & 56.41 & 32.57 & 00.21 & 74.93 & 71.11 & 64.98 & 59.29 \\
AVFusion~\cite{bagchiHear2022} & 65.38 & 23.89 & 00.11 & 62.98 & 59.26 & 54.80 & 52.11 \\
\modelabbr{}~\cite{caiYou2022} & 79.15 & 38.57 & 00.24 & 67.03 & 64.18 & 60.89 & 58.51 \\
TadTR~\cite{liuEndtoEnd2022} & 80.22 & 61.04 & \textbf{05.22} & 72.50 & 72.50 & 70.56 & 69.18 \\
ActionFormer~\cite{zhangActionFormer2022} & 85.23 & 59.05 & 00.93 & 77.23 & 77.23 & 77.19 & 76.93 \\
TriDet~\cite{shiTriDet2023} & 86.33 & 70.23 & 03.05 & 74.47 & 74.47 & 74.46 & 74.45 \\
\hline
\newmodelabbr{}~(ours) & \textbf{96.30} & \textbf{84.96} & 04.44 & \textbf{81.62} & \textbf{80.48} & \textbf{79.40} & \textbf{78.75} \\
\newmodelabbr{}~(ours)~(visual only) & 64.78 & 54.85 & 02.53 & 64.00 & 59.33 & 55.94 & 54.38 \\
\bottomrule[0.4mm]
\end{tabular}}
\label{tab:fullset}
\end{table*}

\begin{table*}[t]
\centering
\caption{\textbf{Temporal forgery localization results on the ``subset'' of the \datasetabbr{} dataset.} The visual-only version of \newmodelabbr{} uses the output from the visual boundary matching layer, illustrating the performance when using only the visual modality.}
\scalebox{1}{
\begin{tabular}{l||ccc|cccc}
\toprule[0.4 mm]
\rowcolor{mygray}\textbf{Method} & \textbf{AP@0.5} & \textbf{AP@0.75} & \textbf{AP@0.95} & \textbf{AR@100} & \textbf{AR@50} & \textbf{AR@20} & \textbf{AR@10} \\ \hline \hline
BMN~\cite{linBMN2019} & 28.10 & 05.47 & 00.01 & 55.49 & 54.44 & 52.14 & 47.72 \\
BMN (E2E) & 32.32 & 11.38 & 00.14 & 59.69 & 48.17 & 39.01 & 34.17 \\
MDS~\cite{chughNot2020} & 23.43 & 03.48 & 00.00 & 58.53 & 56.68 & 53.16 & 49.67 \\
AGT~\cite{nawhalActivity2021} & 15.69 & 10.69 & 00.15 & 49.11 & 40.31 & 31.70 & 23.13 \\
BSN++~\cite{suBSN2021} & 65.26 & 37.70 & 00.22 & 78.89 & 76.32 & 71.00 & 65.38 \\
AVFusion~\cite{bagchiHear2022} & 62.01 & 22.77 & 00.11 & 61.98 & 58.08 & 53.31 & 50.52 \\
\modelabbr{}~\cite{caiYou2022} & 85.20 & 47.06 & 00.29 & 67.34 & 64.52 & 61.19 & 59.32 \\
TadTR~\cite{liuEndtoEnd2022} & 83.48 & 63.57 & \textbf{05.44} & 74.15 & 74.15 & 72.42 & 71.38 \\
ActionFormer~\cite{zhangActionFormer2022} & 79.48 & 48.01 & 01.08 & 70.38 & 70.38 & 70.36 & 70.08 \\
TriDet~\cite{shiTriDet2023} & 80.71 & 60.93 & 02.91 & 67.64 & 67.64 & 67.64 & 67.63 \\
\hline
\newmodelabbr{}~(ours) & \textbf{96.82} & \textbf{86.47} & 03.90 & \textbf{81.74} & \textbf{80.59} & \textbf{79.60} & \textbf{79.15} \\
\newmodelabbr{}~(ours)~(visual only) & 96.47 & 82.02 & 03.79 & 80.65 & 79.00 & 77.46 & 76.90 \\
\bottomrule[0.4mm]
\end{tabular}}
\label{tab:subset}
\end{table*}

\subsubsection{Multimodal Boundary Matching Loss}
We utilized the label information for each modality to train the proposed multimodal framework and extended the concept of boundary matching loss ($\mathcal{L}_b$) to more modalities. The multimodal boundary matching loss is defined as follows,

$$\mathcal{L}_{bm} = \frac{1}{2D\sum\mathbb{T}}\sum_{m\in\{v,a\}}\sum_{\alpha\in\{p,c,pc\}}\sum^n_{i=1}\sum^D_{j=1}\sum^{t_i}_{k=1}(\hat{Y}^{(b)(\alpha)}_{mijk} - Y^{(b)}_{mijk})^2 $$
$$Y^{(b)}_m = \eta_m Y^{(b)} + (1 - \eta_m)Y^{(b)}_\phi,$$
where, $m$ is the modality (visual $v$ or audio $a$), $\eta_m$ specifies whether modality $m$ is modified, $\alpha$ is one of the boundary map types from the boundary module, $Y^{(b)}_\phi \in 0^{D\times T}$ is the ground truth boundary maps for real videos, and $\mathbb{T} = \{t_i\}_0^{n}$ where $\sum\mathbb{T}$ is the total number of frames in the dataset.

\subsubsection{Overall Loss} The overall training objective of \newmodelabbr{} is defined as,

$$\mathcal{L} = \mathcal{L}_b + \lambda_{bm} \mathcal{L}_{bm} + \lambda_f \mathcal{L}_f + \lambda_c \mathcal{L}_c,$$
where, $\lambda_{bm}$, $\lambda_f$ and $\lambda_c$ are weights for different losses.

\begin{table*}[t]
\centering
\caption{\textbf{Temporal forgery localization results on the ForgeryNet dataset.} The visual-only version of \newmodelabbr{} uses the output from the visual boundary matching layer, illustrating the performance when using only the visual modality.}
\scalebox{0.92}{
\begin{tabular}{l||c|ccc|ccc}
\toprule[0.4 mm]
\rowcolor{mygray}\textbf{Method} & \textbf{Avg. AP} & \textbf{AP@0.5} & \textbf{AP@0.75} & \textbf{AP@0.95} & \textbf{AR@5} & \textbf{AR@2} \\ \hline \hline
Xception~\cite{cholletXception2017} & 62.83 & 68.29 & 62.84 & 58.30 & 73.95 & 25.83 \\
X3D-M+BSN~\cite{feichtenhoferX3D2020, linBSN2018} & 70.29 & 80.46 & 77.24 & 55.09 & 86.88 & 81.33 \\
X3D-M+BMN~\cite{feichtenhoferX3D2020, linBMN2019} & 83.47 & 90.65 & 88.12 & 74.95 & 91.99 & 88.44 \\
SlowFast+BSN~\cite{feichtenhoferSlowFast2019, linBSN2018} & 73.42 & 82.25 & 80.11 & 60.66 & 88.78 & 83.63 \\
SlowFast+BMN~\cite{feichtenhoferSlowFast2019, linBMN2019} & 86.85 & 92.76 & \textbf{91.00} & 80.02 & 93.49 & \textbf{90.64} \\
\hline
\newmodelabbr{}~(ours)~(visual only) & \textbf{87.79} & \textbf{93.13} & 89.14 & \textbf{81.09} & \textbf{95.69} & 90.63 \\
\bottomrule[0.4mm]
\end{tabular}}
\label{tab:forgerynet}
\end{table*}

\subsection{Inference}
During inference, the model generates three types of fusion boundary maps - position-aware boundary map $\hat{Y}^{(b)(p)}$, channel-aware boundary map $\hat{Y}^{(b)(c)}$ and aggregated position-channel boundary map $\hat{Y}^{(b)(pc)}$. Following previous work~\cite{suBSN2021}, we averaged the three boundary maps to produce the final boundary map $\hat{Y}^{(b)}$. This boundary map represents the confidence for all proposals in the video. Since this operation produces duplicated proposals, we post-process the proposals with Soft Non-Maximum Suppression (S-NMS)~\cite{bodlaSoftNMS2017} similar to BSN++~\cite{suBSN2021}.

\section{Experiments}
\label{sec:experiments}

\begin{table}[t]
\centering
\caption{\textbf{Deepfake detection results on the DFDC dataset.}}
\scalebox{1}{
\begin{tabular}{l|c}
\toprule[0.4 mm]
\rowcolor{mygray} \textbf{Method} & \textbf{AUC} \\ \hline \hline
Meso4~\cite{afcharMesoNet2018} & 0.753 \\
FWA~\cite{liExposing2019} & 0.727 \\
Siamese~\cite{mittalEmotions2020} & 0.844 \\
MDS~\cite{chughNot2020} & 0.916 \\
\modelabbr{}~\cite{caiYou2022} & 0.846 \\ \hline
\newmodelabbr{}~(ours) & \textbf{0.937} \\
\bottomrule[0.4mm]
\end{tabular}}
\label{tab:dfdc}
\end{table}

\subsection{Dataset Partitioning}
We splitted the \datasetabbr{} dataset into 78,703 train, 31,501 validation and 26,100 test videos. The test partition is denoted as \textit{full set}. For a fair comparison with existing visual-only methods~\cite{linBMN2019, suBSN2021}, we additionally prepared a subset of the full set denoted as \textit{subset} where the audio-only manipulated videos are removed.

\subsection{Implementation Details}
The \newmodelabbr{} method is implemented in PyTorch~\cite{paszkePyTorch2019} and the model is trained using 2 NVIDIA A100 80GB GPUs. We resized the input videos to $96 \times 96$ to reduce the computational cost of the MViTv2-based visual backbone. The temporal dimension $T$ is fixed to 512 for \datasetabbr{} and 300 for ForgeryNet~\cite{heForgeryNet2021} and DFDC~\cite{dolhanskyDeepFake2020}. The latent features $\mathbf{z}_v$ and $\mathbf{z}_a$ have the same shape $C_f \times T$ where the feature size $C_f = 256$ and $T \in \{512, 300\}$. For the boundary matching modules $\mathcal{F}_{\mathcal{B}_v}$ and $\mathcal{F}_{\mathcal{B}_a}$, we set the maximum segment duration $D$ to 40 for \datasetabbr{}, 200 for ForgeryNet and 300 for DFDC. We followed the training protocol proposed in MViTv2~\cite{liMViTv22022}.
Throughout our experiments, we empirically set $\lambda_c$ = $0.1$, $\lambda_f$ = $2$, $\lambda_b$ = $1$, $\lambda_{bm}$ = $1$ and $\delta$ = $0.99$.

\subsection{Evaluation Details}
We benchmarked the \datasetabbr{} dataset for deepfake detection and localization tasks. For deepfake detection we follow standard evaluation protocols~\cite{rosslerFaceForensics2019, dolhanskyDeepFake2020}, and use Area Under the Curve (AUC) as evaluation metric for this binary classification task. We are the first to benchmark deepfake localization task and adopt Average Precision (AP) and Average Recall (AR) as the evaluation metrics. For AP, we set the IoU thresholds to 0.5, 0.75 and 0.95, following ActivityNet~\cite{cabaheilbronActivityNet2015} evaluation protocol. For AR, since the number of fake segments is small, we set the number of proposals to 100, 50, 20 and 10 with the IoU thresholds [0.5:0.05:0.95]. When evaluating the proposed approach on ForgeryNet~\cite{heForgeryNet2021}, we follow the protocol in that paper (i.e. AP@0.5, AP@0.75, AP@0.9, AR@5, and AR@2).

For evaluating \newmodelabbr{} on ForgeryNet, we used only the visual pipeline of the method to train the model (ForgeryNet is a visual-only deepfake dataset). Since only the visual modality is used in the model, only $\mathcal{L}_b$ and $\mathcal{L}_f$ are used for training. Similarly for evaluation on DFDC~\cite{dolhanskyDeepFake2020}, we consider the whole fake video as one fake segment and train our model in the temporal localization manner. Then, we train a small MLP to map the boundary map to the final binary labels.

We also evaluated the performance of several state-of-the-art methods on \datasetabbr{}, including BMN~\cite{linBMN2019}, AGT~\cite{nawhalActivity2021}, AVFusion~\cite{bagchiHear2022}, MDS~\cite{chughNot2020}, BSN++~\cite{suBSN2021}, TadTR~\cite{liuEndtoEnd2022}, ActionFormer~\cite{zhangActionFormer2022}, and TriDet~\cite{shiTriDet2023}. Based on the original implementations, BMN, BSN++, TadTR, ActionFormer, and TriDet require extracted features, thus, we trained these models based on 2-stream I3D features~\cite{carreiraQuo2017}. For the methods that require S-NMS~\cite{bodlaSoftNMS2017} during post-processing, we searched the optimal hyperparameters for S-NMS using the validation part of the concerned dataset. All reported results are based on the test partitions.

\begin{table*}[t]
\centering
\caption{\textbf{Impact of loss functions.} The contribution of different losses for temporal forgery localization on the full set of the \datasetabbr{} dataset.}
\scalebox{1}{
\begin{tabular}{l||ccc|cccc}
\toprule[0.4 mm]
\rowcolor{mygray}\textbf{Loss Function} & \textbf{AP@0.5} & \textbf{AP@0.75} & \textbf{AP@0.95} & \textbf{AR@100} & \textbf{AR@50} & \textbf{AR@20} & \textbf{AR@10} \\ \hline \hline
$\mathcal{L}_f$ & 59.45 & 51.46 & 07.11 & 77.25 & 75.60 & 70.76 & 67.24 \\
$\mathcal{L}_c, \mathcal{L}_f$ & 63.42 & 56.24 & 08.55 & 78.17 & 76.47 & 71.58 & 68.22 \\
$\mathcal{L}_b$ & 71.31 & 34.30 & 00.12 & 66.92 & 63.67 & 57.99 & 54.72 \\
$\mathcal{L}_{bm}, \mathcal{L}_b$ & 71.97 & 51.17 & 00.50 & 69.86 & 67.58 & 64.44 & 62.64 \\
$\mathcal{L}_f, \mathcal{L}_{bm}, \mathcal{L}_b$ & 94.71 & 78.54 & 01.66 & 77.86 & 76.44 & 74.67 & 73.69 \\ \hline
$\mathcal{L}_c, \mathcal{L}_f, \mathcal{L}_{bm}, \mathcal{L}_b$ & \textbf{96.30} & \textbf{84.96} & \textbf{04.44} & \textbf{81.62} & \textbf{80.48} & \textbf{79.40} & \textbf{78.75} \\
\bottomrule[0.4mm]
\end{tabular}}
\label{tab:losses}
\end{table*}

\begin{table*}[t]
\centering
\caption{\textbf{Impact of pre-trained features.} Comparison of different pre-trained features for temporal forgery localization on the full set of the \datasetabbr{} dataset.}
\scalebox{0.85}{
\begin{tabular}{l|l|c||ccc|cccc}
\toprule[0.4 mm]
\rowcolor{mygray}\textbf{Visual} & \textbf{Audio} & \textbf{Citation} & \textbf{AP@0.5} & \textbf{AP@0.75} & \textbf{AP@0.95} & \textbf{AR@100} & \textbf{AR@50} & \textbf{AR@20} & \textbf{AR@10} \\ \hline \hline
I3D & E2E & \cite{carreiraQuo2017} & 74.76 & 59.57 & 04.02 & 74.28 & 71.92 & 68.64 & 66.63 \\
MARLIN & E2E & \cite{caiMARLIN2023} & 92.27 & 75.11 & 04.10 & 77.93 & 76.38 & 74.53 & 73.47 \\
3DMM & E2E & \cite{blanzmorphable1999} & 01.84 & 00.11 & 00.00 & 34.00 & 31.54 & 20.94 & 11.81 \\
E2E	& TRILLsson3 & \cite{shorTRILLsson2022} & 95.16 & 82.67 & 05.65 & 81.21 & 79.80 & 78.22 & 77.49 \\
E2E & Wav2Vec2 & \cite{baevskiwav2vec2020} & 95.92 & 84.94 & \textbf{05.66} & \textbf{82.48} & \textbf{81.38} & \textbf{79.93} & \textbf{79.24} \\ \hline
E2E	& E2E & N/A & \textbf{96.30} & \textbf{84.96} & 04.44 & 81.62 & 80.48 & 79.40 & 78.75 \\
\bottomrule[0.4mm]
\end{tabular}}
\label{tab:features}
\end{table*}

\begin{table*}[!htbp]
\centering
\caption{\textbf{Impact of encoder architectures.} Comparison of different backbone architectures for temporal forgery localization on the full set of the \datasetabbr{} dataset.}
\scalebox{0.95}{
\begin{tabular}{l|l|c||ccc|cccc}
\toprule[0.4 mm]
\rowcolor{mygray}\textbf{Visual} & \textbf{Audio} & \textbf{Boundary} & \textbf{AP@0.5} & \textbf{AP@0.75} & \textbf{AP@0.95} & \textbf{AR@100} & \textbf{AR@50} & \textbf{AR@20} & \textbf{AR@10} \\ \hline \hline
3D CNN & CNN & BMN & 76.90 & 38.50 & 00.25 & 66.90 & 64.08 & 60.77 & 58.42 \\
3D CNN & CNN & BSN++ & 92.44 & 71.34 & 01.15 & 75.86 & 74.43 & 72.39 & 71.21 \\
MViTv2-Tiny & CNN & BMN & 89.32 & 59.47 & 01.45 & 72.52 & 70.14 & 67.55 & 65.92 \\
MViTv2-Small & CNN & BMN & 89.31 & 59.97 & 01.78 & 72.74 & 70.35 & 67.56 & 65.87 \\
MViTv2-Base & CNN & BMN & 89.90 & 59.67 & 01.51 & 72.22 & 69.99 & 67.29 & 65.64 \\
3D CNN & ViT-Tiny & BMN & 78.08 & 35.18 & 00.41 & 67.38 & 64.38 & 60.92 & 58.66 \\
3D CNN & ViT-Small & BMN & 79.61 & 37.63 & 00.42 & 67.10 & 64.23 & 60.77 & 58.51 \\
3D CNN & ViT-Base & BMN & 80.86 & 36.55 & 00.34 & 67.24 & 64.27 & 60.86 & 58.46 \\
MViTv2-Small & ViT-Base & BSN++ & 93.59 & 75.22 & 02.56 & 77.73 & 76.08 & 74.07 & 72.93 \\
MViTv2-Base & ViT-Base & BSN++ & \textbf{96.30} & \textbf{84.96} & \textbf{04.44} & \textbf{81.62} & \textbf{80.48} & \textbf{79.40} & \textbf{78.75} \\
\bottomrule[0.4mm]
\end{tabular}}
\label{tab:modules}
\end{table*}

\begin{figure*}[t]
\centering
\includegraphics[width=\textwidth]{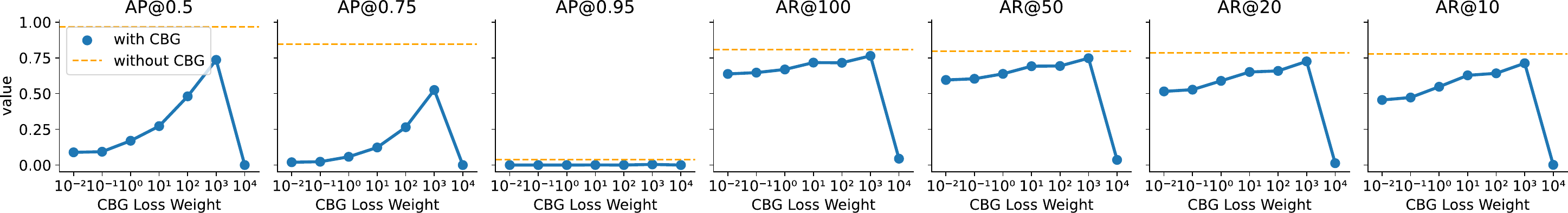}
\caption{\textbf{Impact of CBG in the boundary matching module.} The figure shows comparison of models containing CBG module and a model without CBG module.}
\label{fig:cbg}
\end{figure*}

\section{Results}
\subsection{Temporal Forgery Localization}
\subsubsection{LAV-DF Dataset}
We evaluated the performance of \newmodelabbr{} on the \datasetabbr{} dataset for temporal forgery localization, and compare it with other approaches. For the full set, from Table~\ref{tab:fullset}, our method achieves the best performance for AP@0.5 and AR@100.
Unlike temporal action localization datasets, the segments in our dataset have a single label for the fake segments which leads to high AP scores. The multimodal MDS method is not specifically designed for temporal forgery localization tasks and can only predict fixed-length segments, lacking the ability to precisely identify boundaries. Therefore, the scores for MDS are relatively low.
For BMN and BSN++, the AP scores are low because they are designed for fake proposal generation instead of forgery localization. TadTR, ActionFormer, and TriDet achieve relatively better performance as they are one-stage temporal action localization approaches that generate more precise segments. Additionally, we observe that BMN trained with an end-to-end visual encoder performs better than using pre-trained I3D features. With the multimodal complimentary information, our approach outperforms the aforementioned approaches.

We further evaluated all methods on the subset of the \datasetabbr{} dataset. From Table~\ref{tab:subset}, it is observed that the performance of the visual-only methods including BMN, AGT, BSN++ and TadTR is improved. The visual-only score of our method improves from 64.78~(AP@0.5) to 96.47~(AP@0.5), and the margin between the unimodal and multimodal versions is decreased significantly from 31.52~(AP@0.5) to 0.35~(AP@0.5). Thus, our method demonstrates its superior performance for temporal forgery localization.

\subsubsection{ForgeryNet Dataset}
We evaluated the performance of the visual-only \newmodelabbr{} trained on the ForgeryNet dataset, and compare it with other approaches (using the results reported by~\cite{heForgeryNet2021}). As shown in Table~\ref{tab:forgerynet}, the performance of the visual-only \newmodelabbr{} exceeds the previous best model SlowFast~\cite{feichtenhoferSlowFast2019}+BMN~\cite{linBMN2019}, showing that proposed method has advantage for temporal forgery localization.

\subsection{Deepfake Detection}
We also compare our method with previous deepfake detection methods on a subset of the DFDC dataset following the configuration of~\cite{chughNot2020}. As shown in Table~\ref{tab:dfdc}, the performance of our method is better than previous methods such as Meso4~\cite{afcharMesoNet2018}, FWA~\cite{liExposing2019}, Siamese~\cite{mittalEmotions2020}, and MDS~\cite{chughNot2020}.
In summary, our method performs well on the classification task.

\subsection{Ablation Studies}
\subsubsection{Impact of Loss Functions}
To examine the contributions of each loss of \newmodelabbr{}, we train six models with different combinations of losses. To aggregate the frame-level predictions for the models without boundary module, we follow the algorithm proposed in previous work~\cite{zhaoTemporal2017}. From~\cref{tab:losses}, it is evident that all of the integrated losses have positive influence on the performance. By observing the difference between the scores, the boundary matching loss $\mathcal{L}_b$ and the frame classification loss $\mathcal{L}_c$ contribute significantly to the performance. With the frame-level labels supervising the model, the encoders are trained to have a better capacity to extract the features relevant to deepfake artifacts. Whereas the boundary module mechanism have localization ability to detect the fake segments more precisely.

\subsubsection{Impact of Pre-Trained Features}
In the literature~\cite{liuEndtoEnd2022, zhangActionFormer2022}, pre-trained visual features, such as I3D~\cite{carreiraQuo2017}, are commonly used for temporal action localization. Since the I3D features are pre-trained on the Kinetics dataset~\cite{kayKinetics2017}, they encode the representation of the universal scene of the video. However, temporal forgery localization requires the model to have a specialized understanding of facial information. Therefore, the pre-trained features obtained from universal visual dataset are not likely to be suitable for our task. Our quantitative results support this, e.g. the comparison between the two BMN models in Table~\ref{tab:fullset} where one uses I3D features and the other uses end-to-end training.

To examine the impact of pre-trained features on \newmodelabbr{}, we trained models using different pre-trained features, including visual (I3D, MARLIN ViT-S~\cite{caiMARLIN2023} and 3DMM~\cite{blanzmorphable1999}) and audio features (TRILLsson~\cite{shorTRILLsson2022} and Wav2Vec2~\cite{baevskiwav2vec2020}). The results are shown in Table~\ref{tab:features}. From the results, we can observe the following patterns: 1) The model trained fully end-to-end reaches the best performance and 2) Compared with visual features, audio features have better task specific performance.

\subsubsection{Impact of Encoder Architectures}
To find the best modality-specific architecture for \newmodelabbr{}, we trained several architecture combinations for the visual encoder, audio encoder, and boundary module. The results are presented in Table~\ref{tab:modules}. Compared to the previous model \modelabbr{}~\cite{caiYou2022} as baseline (3D-CNN + CNN + BMN~\cite{linBMN2019}), we used the attention-based architectures including MViTv2~\cite{liMViTv22022} and ViT~\cite{dosovitskiyImage2021} families for encoders and attention-based BSN++ modules~\cite{suBSN2021} for predicting boundaries. 

We used the variations of MViTv2 from the original paper (\ie MViTv2-Tiny, MViTv2-Small and MViTv2-Base) as the visual encoders. We can conclude that the MViTv2 architecture plays an important role while comparing with the baseline, but the benefit of different scales of the MViTv2 architecture is not significant. As for the audio encoder, we followed the architecture definitions for ViT (\ie ViT-Tiny, ViT-Small and ViT-Base) for comparison. We can conclude that the audio encoder benefits from different scales of the ViT architecture. We also compared the BSN++-based boundary module with BMN-based architecture. The contribution from the BSN++ is the most significant compared with MViTv2 for the visual encoder and ViT for the audio encoder. Owing to the attention mechanism, the framework utilizes the global and local context to analyze the artifacts. The combination of MViTv2-Base, ViT-Base and BSN++ produces the best performance compared to all other combinations of modules.

\subsubsection{Impact of CBG in the Boundary Matching Module}
\label{sec:cbg}
We adopted the method from BSN++~\cite{suBSN2021} to improve the performance for temporal forgery localization. This method includes two modules, complementary boundary generator (CBG) and proposal relation block (PRB). The CBG module predicts the confidence that a timestamp is starting or ending point of segments. The PRB module, based on BMN~\cite{linBMN2019}, predicts the boundary map which contains the confidences of dense segment proposals. For inference, the results from both modules are multiplied as the final output. In this ablation study, we aim to discuss the impact of the CBG module.

We trained several models containing CBG modules with different loss weights, from $10^{-2}$ to $10^4$, and also a model without CBG module. As shown in Figure~\ref{fig:cbg}, the best CBG loss weight is $10^3$. However, compared with the non-CBG model, the best model with CBG can only compete on AR and has a huge gap on AP metrics. Based on this observation, we drop the CBG module in the boundary module and only use PRB.

\section{Conclusion}
In this paper, we introduce and investigate content-driven multimodal deepfake generation, detection, and localization. We introduce a new dataset where both the audio and visual modalities are modified at strategic locations. Additionally, we propose a new method for temporal forgery localization. Through extensive experiments, we demonstrate that our method outperforms existing state-of-the-art techniques.

The proposed dataset, \datasetabbr{}, may raise ethical concerns due to its potential negative social impact. Given that the dataset contains facial videos of celebrities, there could be a risk of its misuse for unethical purposes. Moreover, the dataset generation pipeline itself can be used to generate fake videos. To confront the potential negative impact of our work, we have taken several measures. Most importantly, we have prepared an end-user license agreement as a preventive measure. Similarly, users need to agree on terms and conditions to use the proposed temporal forgery localization method \newmodelabbr{}.

This work has some limitations: 1) The audio reenactment method employed for dataset creation does not consistently generate the desired reference style, 2) The resolution of the dataset is limited by the source videos, and 3) The high classification scores obtained indicate the need for further improvement in the visual reenactment method.

Major improvement in the future will be extending the generation pipeline to include word tokens insertion, substitution and deletion and converting statements into questions.

\bibliographystyle{ieee_fullname}
\bibliography{refs}

\end{document}